%% file: main.tex
\documentclass[10pt,twocolumn,letterpaper]{article}

\usepackage[pagenumbers]{wacv} % To force page numbers, e.g. for an arXiv version


\definecolor{wacvblue}{rgb}{0.21,0.49,0.74}
\usepackage[pagebackref,breaklinks,colorlinks,allcolors=wacvblue]{hyperref}

\def\wacvPaperID{*****} % *** Enter the WACV Paper ID here
\def\confName{WACV}
\def\confYear{2026}

\title{Real-Time Fusion of Visual and Chart Data for Enhanced Maritime Vision}

\author{Marten Kreis\\
University of Tuebingen\\
{\tt\small marten.kreis@student.uni-tuebingen.de}
\and
Benjamin Kiefer\\
LOOKOUT\\
{\tt\small benjamin@lookout.team}
}

\begin{document}
\maketitle
\input{sec/0_abstract}
\input{sec/1_intro}

\input{sec/2_background}

\input{sec/3_method}
\input{sec/4_experiments}

\input{sec/5_discussion}
{
    \small
    \bibliographystyle{ieeenat_fullname}
    \bibliography{main}
}

\end{document}

%% file: sec/0_abstract.tex
\begin{abstract}
    This paper presents a novel approach to enhancing marine vision by fusing real-time visual data with chart information. Our system overlays nautical chart data onto live video feeds by accurately matching detected navigational aids, such as buoys, with their corresponding representations in chart data. To achieve robust association, we introduce a transformer-based end-to-end neural network that predicts bounding boxes and confidence scores for buoy queries, enabling the direct matching of image-domain detections with world-space chart markers. The proposed method is compared against baseline approaches, including a ray-casting model that estimates buoy positions via camera projection and a YOLOv7-based network extended with a distance estimation module. Experimental results on a dataset of real-world maritime scenes demonstrate that our approach significantly improves object localization and association accuracy in dynamic and challenging environments.
\end{abstract}

%% file: sec/1_intro.tex
\section{Introduction}
\label{sec:intro}

Accurate navigation in maritime environments is critical yet challenging, especially when conventional navigational sensors such as GPS, satellite compasses, and IMUs lack the precision needed in dynamic conditions. Visual data, captured by onboard cameras, provides rich spatial information that can be leveraged to enhance situational awareness. However, integrating this visual information with existing chart data to yield precise, real-time guidance remains an open problem. 

\begin{figure}
    \centering
    \includegraphics[width=0.98\linewidth]{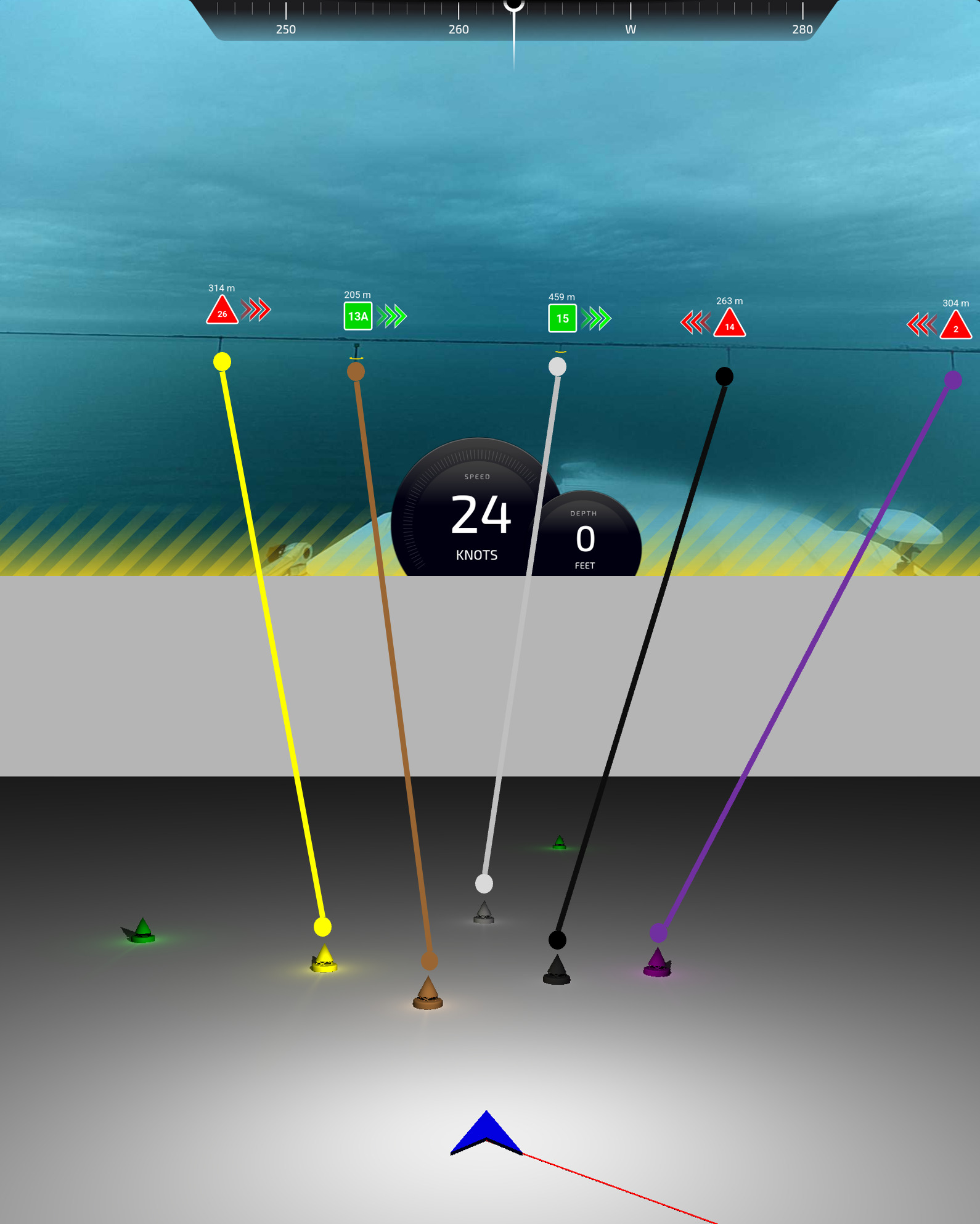}
    \caption{AR view overlaying nautical chart data onto the video frame, leveraging the computed correspondence between mapped buoys and detected objects in the image domain using the Fusion Transformer.}
    \label{fig:intro_pic}
\end{figure}

In this work, we tackle the problem of matching objects detected in live video feeds with their corresponding navigational aids in nautical charts. Our focus is on the detection and precise localization of buoys, which serve as crucial navigational markers in marine environments. By overlaying chart data onto video streams, our system assists operators in navigating complex waterways by providing both the identity and the direction of nearby navigational aids. This is depicted in Figure \ref{fig:intro_pic}.

To bridge the gap between visual and chart domains, we propose a system that integrates multiple computer vision techniques with robust data association methods. At the core of our approach is a transformer-based end-to-end neural network that predicts bounding boxes and associated confidence scores for buoy queries sampled from chart data. This method is designed to overcome the limitations of traditional projection-based techniques, such as those relying solely on ray-casting from camera coordinates to the world frame, and to improve upon YOLO-based detectors augmented with distance estimation.

% \begin{figure}
%   \centering
%   \begin{subfigure}{1\linewidth}
%     \includegraphics[width=\linewidth]{img/AR_View.png}
%     \caption{AR Application with visual overlay of nautical chart data.}
%     \label{fig:short-a}
%   \end{subfigure}
%   \hfill
%   \begin{subfigure}{1\linewidth}
%     \includegraphics[width=\linewidth]{img/intro_association.png}
%     \caption{Association between bounding boxes and mapped channel markers, computed by the Fusion Transformer. Green buoys are sampled unmatched queries, i.e. they are not visible in the frame}
%     \label{fig:short-b}
%   \end{subfigure}
%   \caption{Inference results of Fusion Transformer: Correspondence between chart markers and detected objects in the image. }
%   \label{fig:short}
% \end{figure}

The contributions of this paper are threefold:
\begin{enumerate}
    \item We propose a new fusion framework that overlays chart data onto live video by matching detected objects with corresponding navigational markers in an end-to-end approach.
    \item We develop a transformer-based network that jointly learns to predict object locations and perform the matching task, thereby enhancing the accuracy of buoy association.
    \item We provide a comprehensive experimental evaluation on real-world maritime datasets, including the development of a labeling tool to generate ground truth for matching accuracy. We're releasing all code, captured videos, sourced chart data and corresponding labels plus labeling tool for easy reproduction of our results.
\end{enumerate}

The remainder of the paper is organized as follows. Section~\ref{sec_background} reviews related work in object detection, tracking, and sensor fusion for marine navigation. Section~\ref{sec_method} details the proposed system architecture of the Fusion Transformer. Section~\ref{sec_experiments} presents our experimental setup and results, whilst Section~\ref{sec_discussion} concludes the paper with discussions on future work.

%% file: sec/2_background.tex
\section{Related Work}
\label{sec_background}

%\subsection{Maritime Object Detection and Tracking}
\subsection{Maritime Object Detection}

Maritime computer vision has advanced with specialized datasets and methods for open water object detection. The SeaDronesSee benchmark~\cite{varga2022seadronessee} supports human detection in maritime environments, while Poseidon~\cite{ruiz2023poseidon} provides augmentation techniques to improve small object detection.

The MaCVi initiative~\cite{kiefer20231st, kiefer20232nd,kiefer20253rd} has been central to benchmarking vision algorithms for USVs and UAVs, with methods such as real-time horizon locking~\cite{kiefer2024real}, stable yaw estimation~\cite{kiefer2023stable}, and memory maps for robust video object detection~\cite{kiefer2023memory}. Benchmarks like MODS~\cite{bovcon2021mods} and MaSTr1325~\cite{bovcon2019mastr1325} have contributed to maritime obstacle detection, while the MARVEL dataset~\cite{gundogdu2017marvel} and maritime-specific benchmarks~\cite{moosbauer2019benchmark} focus on vessel detection. Advanced techniques such as IMU-assisted stereo vision~\cite{bovcon2018stereo} and fast image-based detection methods~\cite{kristan2015fast} further improve segmentation accuracy.

%For tracking, various approaches have been explored: ship classification methods~\cite{leclerc2018ship}, video-based vessel tracking~\cite{moreira2014survey}, saliency-based tracking~\cite{cane2016saliency}, and person-overboard detection strategies~\cite{hoehner2024object}. Additionally, Ahmed et al.~\cite{ahmed2024seal} introduced the Seal Pipeline to enhance dynamic object tracking for USVs.

\subsection{Distance Estimation Techniques}

Accurate distance estimation is essential for maritime navigation and collision avoidance. Traditional radar methods, improved by modeling echoes as ellipses, provide robust range measurements for large vessels~\cite{czaplewski2021method}. LiDAR systems, while offering high-resolution range data, face cost and environmental constraints~\cite{li2020lidar}. SONAR is effective for underwater detection but less suitable for surface obstacle localization~\cite{bae2015development}. 

AIS data can be exploited to infer inter-vessel distances, improving situational awareness~\cite{tu2017exploiting}. Vision-based methods, including monocular video techniques~\cite{gladstone2016distance} and supervised object distance estimation for USVs~\cite{kiefer2025approximate}, offer cost-efficient alternatives. Sensor fusion combining these modalities further enhances distance estimation accuracy~\cite{thombre2020sensors}.

\subsection{Fusion with Geospatial Data}
\label{sec:geofusion}

In terrestrial autonomous driving, sensor fusion with high-definition maps is common. However, maritime geofusion, particularly for buoy localization, poses unique challenges: the water surface is dynamic, persistent landmarks are scarce, and AIS updates are intermittent. While GIS-assisted localization has been effective in structured environments~\cite{ardeshir2014gis}, and underwater SLAM methods offer promise~\cite{aitken2021simultaneous}, their adaptation to buoy detection must account for environmental variability.

Recent work has combined image and AIS data for maritime applications~\cite{Gulsoylu_2024_WACV}. Additionally, studies on automatic geotagging from street view imagery~\cite{krylov2018automatic, biljecki2021street} illustrate both the potential and limitations of fusing visual detections with nautical charts in unstructured maritime environments.

\subsection{Transformer-Based Object Detection}
DEtection TRansformers (DETR), introduced by Carion \textit{et al.} \cite{carion_end--end_2020}, enable end-to-end object detection using the Transformer architecture \cite{10.5555/3295222.3295349}. This approach eliminates the need for manually designed anchors or non-maximum suppression (NMS). Its simplicity has garnered significant interest within the object detection research community, leading to various improvements aimed at accelerating training convergence, enhancing detection performance, and reducing computational costs \cite{li2022dn, liu2022dabdetr, zhang2022dino, Dai_2021_ICCV, Dai_2021_CVPR, Meng_2021_ICCV, Chen_2023_ICCV, yao2021efficientdetrimprovingendtoend}. Notably, Deformable-DETR \cite{zhu_deformable_2021} optimizes the attention mechanism by focusing on relevant embeddings rather than attending to the entire sequence, improving efficiency. Real-Time DETR \cite{lv2023detrs} achieves performance comparable to state-of-the-art (SOTA) object detectors from the YOLO family while enabling real-time processing.
% Transformers are emerging for maritime perception tasks, excelling in long-range dependencies and robust object association. DETR \cite{carion2020end} has been adapted for maritime scenarios, and Makhmudov \textit{et al.} \cite{makhmudov2023fire} demonstrated superior performance over CNNs. Hybrid models like TR-YOLO \cite{liu2023tr} integrate transformers for improved detection of small objects.

%% file: sec/3_method.tex
\section{Method}
\label{sec_method}

 \begin{figure*}[ht!]
     \centering
     \includegraphics[width=0.95\linewidth]{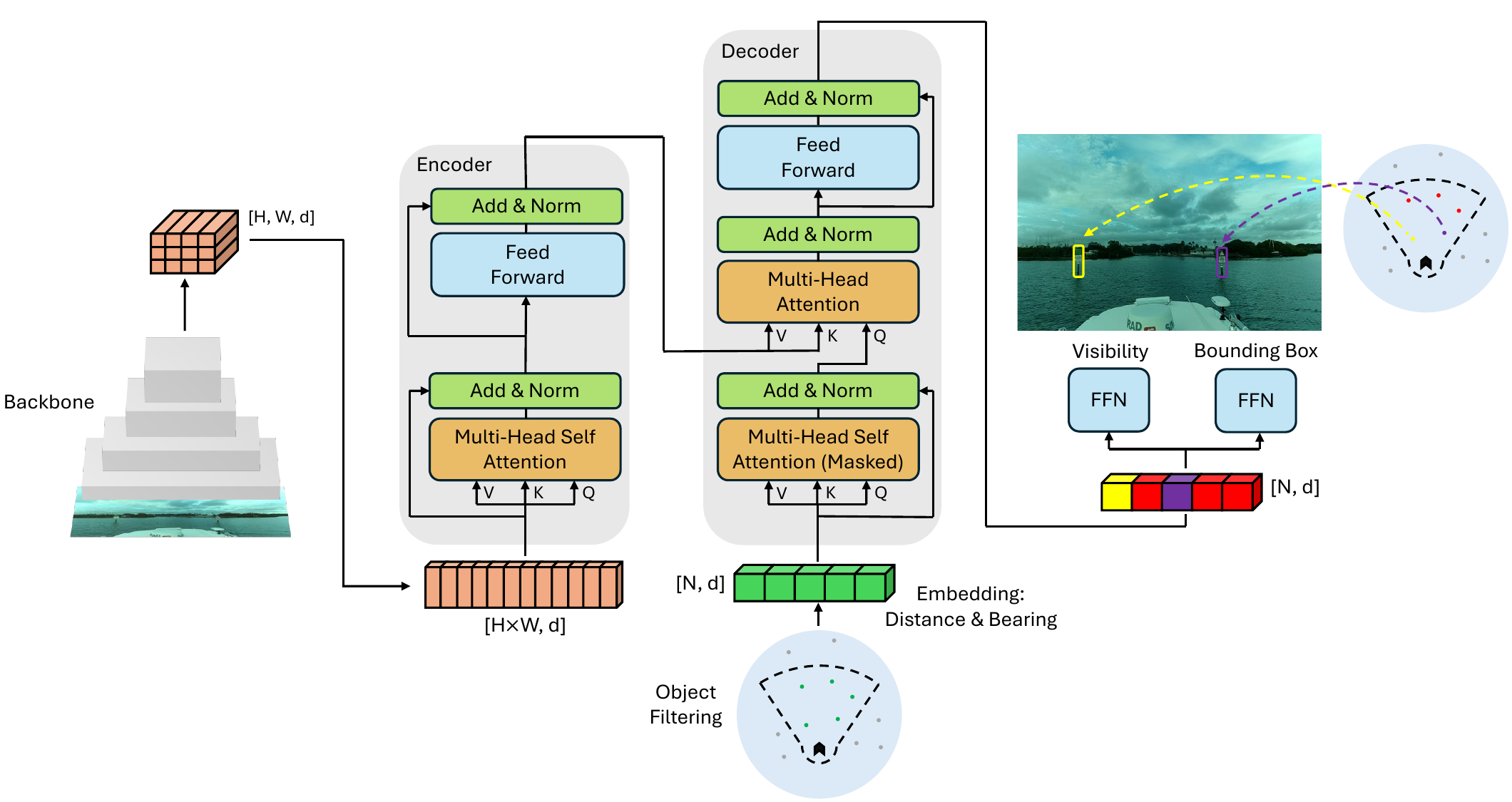}
     \caption{Architecture of the DETR-based Fusion Transformer. A variable number of buoy queries are sampled from nautical chart that lie within the specified FoV. They are used as input embeddings for the Decoder. Each buoy query predicts its visibility for the given frame alongside bounding box coordinates.}
     \label{fig:architecture_fnn}
 \end{figure*}
\subsection{End-to-End Fusion Transformer}
To perform End-to-End matching between detected objects in the image and nearby chart markers, we modify the DETR model \cite{carion_end--end_2020}. Specifically, we incorporate a selected set of buoys -- filtered based on their relative position to the camera -- as input to the decoder. The decoder input embeddings $\mathbf{v}\in \mathbb{R}^{N \times d_{model}}$ are obtained by passing the selected features of each object, for instance the distance and relative bearing to the camera, through a Multi Layer Perceptron (MLP). The Decoder performs Multi-Head Self attention between the buoy queries and Encoder Decoder cross attention between the extracted image features and the buoy queries. The computed output embeddings are passed to a Feed Forward Network (FFN) to obtain the bounding box predictions and an estimate whether the object is visible in the current frame. This approach is depicted in Figure \ref{fig:architecture_fnn}.

\subsubsection{Chart Marker Selection}
\label{cms}
The first step in our proposed approach involves preprocessing the decoder inputs by identifying relevant objects from the chart that could potentially appear in the frame. The navigational aids are filtered, retaining only those objects that either lie within the camera's field of view (FoV) and do not exceed a maximum distance threshold d\_max, or are situated close enough to the camera, within the minimum distance threshold d\_min. The set $\mathcal{S}$ consists of buoys located on the same tile as the camera, which are queried from the NOAA database for maritime navigational aids \cite{noaa_db}. Each entry in the database contains a tuple with the geodetic coordinates (\textit{latitude}, \textit{longitude}) of the object, as well as additional attributes, including a description, category, and ID. We transform the buoys $b_i\in \mathcal{S}$ into the ship's body coordinate system, obtaining coordinates $x_b,y_b,z_b$. Since this method is developed for maritime applications, we assume $z_b=0$, as both the ship and the buoy are situated on the same $xy$-plane. The threshold and FoV parameters are set conservatively to ensure that more buoys are selected than are likely visible. This approach minimizes the risk of inadvertently omitting visible buoys during object selection, which could result from inaccuracies in the vessels estimated position and heading or errors in the chart data.

% \begin{algorithm}
% \caption{Object Selection}
% \label{alg:obj_selection}
% \begin{algorithmic}
%     \State \textbf{Input:} \\
%     Set of buoys sampled from chart data $\mathcal{S} = \{(lat,lng) \:| \:lat\in [-90,90), lng\in [-180,180)\}$ consisting of tuples containing geodetic coordinates
%     \State \textbf{Output:} Subset $\mathcal{S'} \subseteq \mathcal{S}$ containing filtered buoys w.r.t. given thresholds
%     \Statex  % Adds a blank line for readability
%     \State
%     $S' = \emptyset$;\\
%     fov\_with\_padding = 135; \Comment{Degrees} \\
%     dist\_max = 1000; \Comment{Meters}\\
%     dist\_min = 50; \Comment{Meters}
%     \For{each buoy $b\in \mathcal{S}$}
%         \State $x,y,z$ = LatLng2ECEF($b$.lat, $b$.lng);
%         \State $[x_b,y_b,z_b, 1]^T$ = $^{Body} \mathbf{T}_{ECEF} \cdot [x, y, z, 1]^T$;
%         \State $bearing = arctan2(y_b,x_b) \cdot \frac{180}{\pi}$;
%         \State $dist = \sqrt{x_b^2+y_b^2}$;
%         \If{$|bearing| <= \text{fov\_with\_padding} / 2$ and $dist < \text{dist\_max}$}
%             \State $S'$.add($b$);
%         \ElsIf{$dist<\text{dist\_min}$}
%             \State $S'$.add($b$);
%         \EndIf
%     \EndFor
%     \State \Return $S'$
% \end{algorithmic}
% \end{algorithm}

\subsubsection{Model Architecture}
\label{sec:model_architecture}

\begin{figure*}[ht!]
     \centering
     \includegraphics[width=1\linewidth]{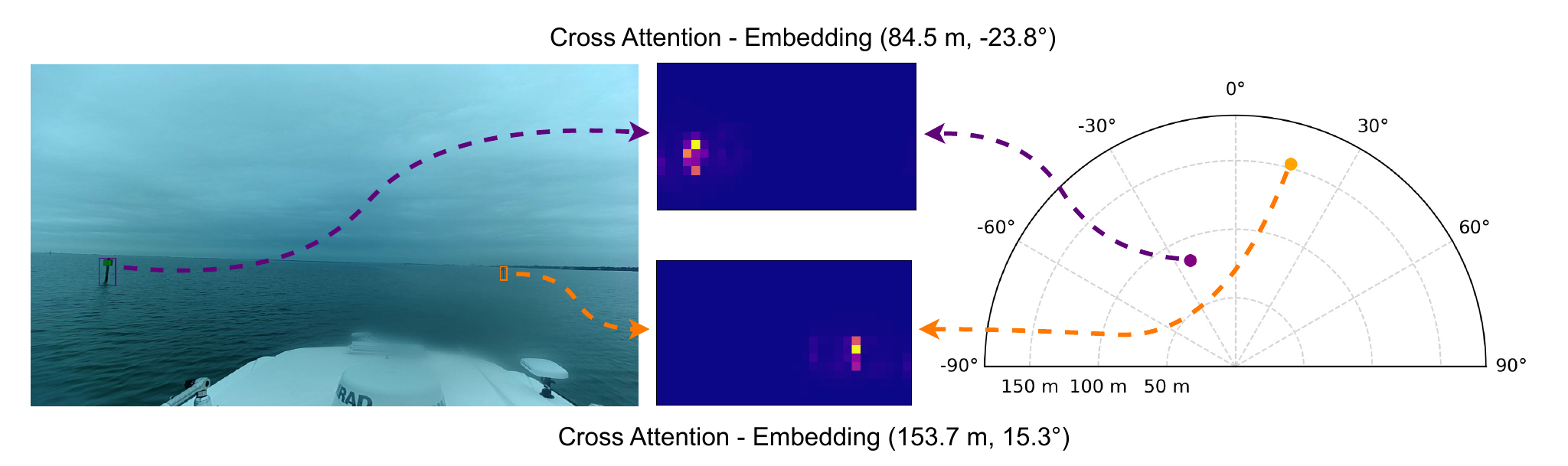}
     \caption{Cross-attention maps for a representative sample, computed using the proposed model. The spatially encoded object information guides the attention mechanism, linking decoder embeddings to relevant visual features in the encoder output.}
     \label{fig:attn_plots}
\end{figure*}

The feature extraction module, consisting of backbone and Encoder is adopted directly from the DETR model \cite{carion_end--end_2020} and not changed. 
% Here, the input image $x_{img}\in \mathbb{R}^{3 \times 540 \times 960}$ is first processed by multiple convolutional and pooling layers that are part of the backbone. A feature map of size $x_{fmap} \in \mathbb{R}^{C \times 17\times30}$ is obtained. A 1x1 convolution reduces the channel dimensions $C$ of the high level feature map to the model dimension $d_{model}$. For the channel dimension we use $C=2048$ and $d_{model}$ is typically set to 256. Furthermore, the embedding of the feature map is flattened, resulting in a vector of the following shape [$H\times W$, $d_{model}$]. A fixed 2D sinusoidal positional encoding is added to the input embedding to preserve spatial information. The encoder extracts high level semantic information, including separating and segmenting different objects. The processes embeddings are passed to the cross attention modules of the Decoder.\\
Instead of passing a fixed number of object queries to the Decoder as input, where each query represents a possible object in the frame, we exchange these learned embeddings with a varying number of selected objects from the chart. These queries can be interpreted as a strong prior $p$, which encodes a spatial bias toward regions in the image where an object is likely to be located. This is depicted in Figure \ref{fig:attn_plots}. Formally, each query $q_i$ is associated with a position $p_i \in [0,1]^2$ in the normalized image space, providing an initial estimate for the object's location. The set $S'$ of filtered objects is obtained as detailed in \ref{cms} and consists of a number of selected features for each object. In our case, the Set $S' = \{(dist,bearing) \:| \: dist\in \mathbb{R}^+, bearing \in[-\pi, \pi)\}$ consists of tuples containing bearing and distance between object and camera. Since the transformer Decoder module requires input embeddings of size $d_{model}$ we employ an MLP, consisting of input layer, one hidden layer and output layer. This yields $N$ embedding vectors $\mathbf{x}_{buoy} \in \mathbb{R}^{d_{model}}$, where $N$ is the number of filtered objects. Note that this number may vary depending on the density of surrounding buoys. Furthermore, we do not apply positional encoding to the object queries, as permutations in the query sequence should not affect the output predictions. By design, the transformer architecture is inherently equivariant to permutations in the input sequences, hence applying positional encoding is not desired.\\
To facilitate batched training, a vector $\mathbf{x}_{queries,batched} \in \mathbb{R}^{b \times N_{max} \times d_{model}}$ is created, where the vectors of individual instances $\mathbf{x}_{queries,i} \in \mathbb{R}^{N_i \times d_{model}}$ are zero padded with the difference between $N_i$ and $N_{max}$. Here, $N_{max}$ represents the longest sequence length over all selected samples. To avoid valid object queries attending to padded queries, masks are applied in the self attention layer, ensuring that the attention weights are set to zero for non valid instances. The predicted outputs for padded queries are not considered during backpropagation and loss computation.\\
After the embeddings are passed through the Decoder, a FFN in conjunction with a sigmoid activation function is employed to predict normalized bounding box coordinates in the format [$c_x$, $c_y$, $w$, $h$] and a visibility score $v \in [0,1]$ for each buoy query, which represents the predicted probability that the object is visible in the frame. Since each query is directly associated with a chart marker, the network's output predictions fuse information from both the chart data and the image domain, thereby establishing correspondence between detected objects and their respective location.

\begin{figure}[ht!]
    \centering
    \includegraphics[width=0.95\linewidth]{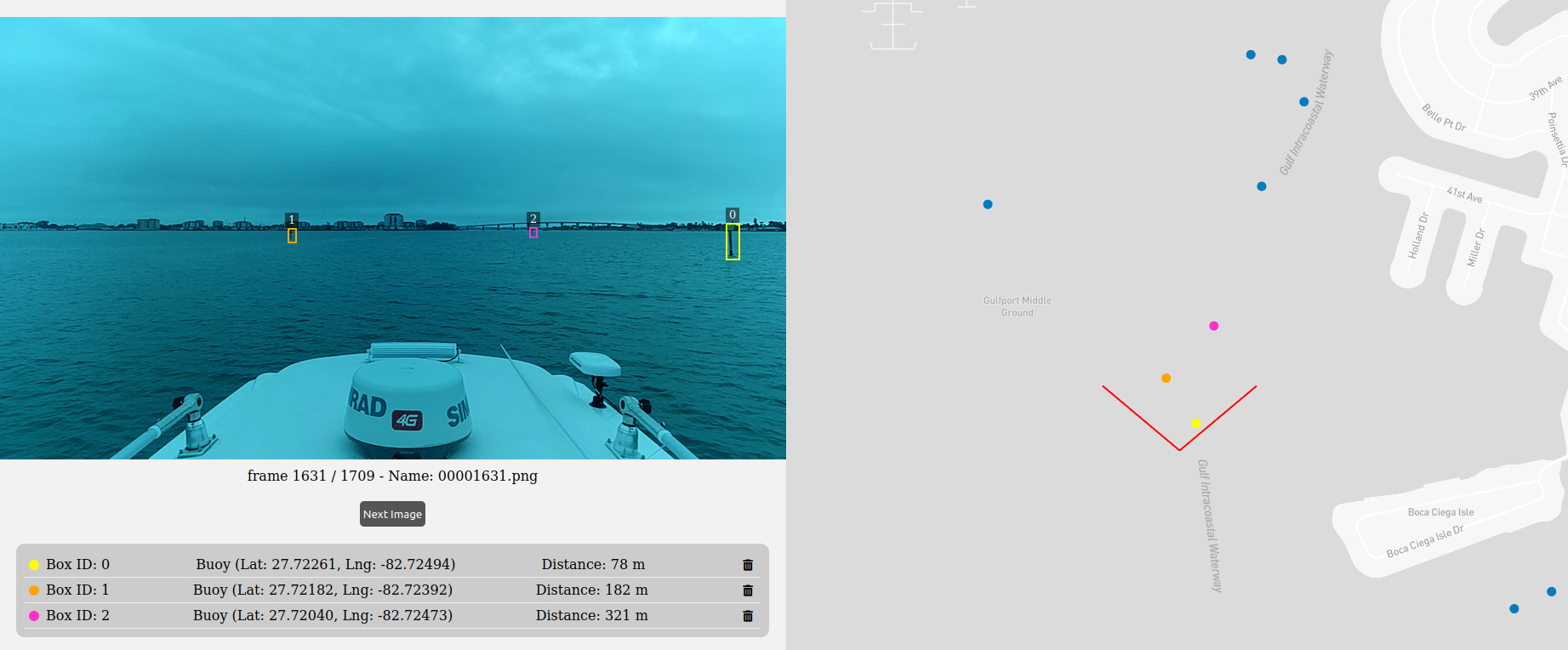}
    \caption{A custom-built labeling tool designed to establish correspondence between bounding box coordinates and chart markers.}
    \label{fig:labeling_tool}
\end{figure}

\subsubsection{Loss Function}
Contrary to DETR based object detectors our approach does not require to compute a hungarian matching between predictions and labels, since the correspondence between ground truth bounding box labels and queries is known. The loss function $\mathcal{L}(y, \hat{y})$ is defined as follows:
\begin{align*}
     \sum_{i=1}^N [v_i\log(\hat{v}_i) + (1-v_i)\log(1-\hat{v}_i)] + v_i \cdot\mathcal{L}_{box}(b_i, \hat{b}_i),
\end{align*}
where $y$ and $\hat{y}$ are gt label and prediction respectively, $v_i, \hat{v}_i$ the visibility scores and $b_i, \hat{b}_i$ the bounding box coordinates. The loss for the visibility prediction $\hat{v_i}$ is computed using Binary Cross Entropy Loss (BCE). In practice, we divide the visibility loss by the number of queries in the batch. The box loss is derived directly from DETR, combining L1 loss and GIoU Loss \cite{carion_end--end_2020}. It is only computed for queries that have a corresponding ground truth label, i.e. where the visibility score $v_i=1$.

\subsection{Dataset}
\begin{figure}
    \centering
    \includegraphics[width=0.99\linewidth]{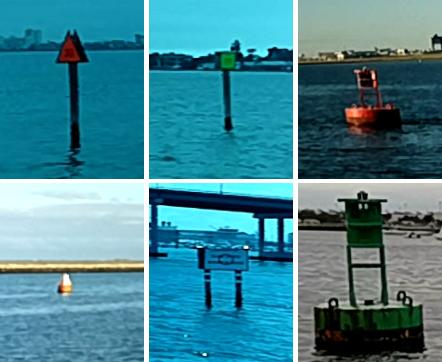}
    \caption{Examples of navigational aids occuring in the dataset are of different shape and color.}
    \label{fig:dataset_buoy_types}
\end{figure}

To train and evaluate the proposed model, we first construct a dataset in which each labeled object within the image domain is matched to its corresponding chart marker. The dataset will be made publicly available. Buoys and other navigational aids that are absent from the chart data are not labeled and are instead treated as negatives. The video data is collected using rented boats operating along the east coast of the United States. The position of the vessel to which the camera is mounted is continuously tracked and recorded. From these videos, individual frames are extracted, ensuring a diverse selection of scenes that capture a range of challenging maritime scenarios. Each frame is annotated with the present navigational aids, see Figure \ref{fig:dataset_buoy_types} for examples.
To establish a reliable correspondence between ground truth bounding boxes in the images and their respective chart markers, we develop a custom web-based labeling tool, as illustrated in Figure \ref{fig:labeling_tool}. This tool enables users to select buoys directly on the chart, which are sampled from the NOAA database for maritime navigational aids \cite{noaa_db}, and associate them with the previously labeled bounding boxes. For generating bounding box annotations, any labeling tool that produces outputs in accordance with the YOLO format convention can be utilized.
The dataset is split into training and validation subsets, consisting of 6,257 and 1,052 samples, respectively, covering a total of 10,977 distinct navigational aids of varying shapes and colors. These instances are extracted from 47 different video sequences. To evaluate the performance of different models, we employ a dedicated test set comprising frames from a single video sequence, where each frame is exhaustively labeled. This setup maintains temporal consistency across frames, which is essential for the developed benchmark models. 
%The approach tracks bounding boxes over consecutive frames and determines a matching confidence based on the duration of uninterrupted association, ensuring that a buoy remains consistently matched without interference from other objects.
%The most common navigation aids that occur in the data set are visualized in Figure \ref{fig:buoy_samples}. 

%\begin{figure}[ht!]
%    \centering
%    \includegraphics[width=0.6\linewidth]{img/buoy_samples.jpg}
%    \caption{Illustration of the most common buoy types in the dataset}
%    \label{fig:buoy_samples}
%\end{figure}

%% file: sec/4_experiments.tex
\section{Experiments}
\label{sec_experiments}
In this section, the performance of the proposed fusion transformer is evaluated against a basic baseline model that relies on ray casting, as well as a distance estimation framework, based on YOLOv7 \cite{wang2023yolov7}, which is combined with the Hungarian matching algorithm. Our experiments assess both the accuracy of the predicted correspondences between the image domain and the chart, alongside latency measurements. Additionally, we conduct ablation studies to examine the impact of architectural modifications to the fusion transformer network.

\subsection{Training and Inference}
The models are exclusively trained on the train split of the created dataset. For the transformer network a sample consists of the following components:
\begin{itemize}
    \item Image, resized to $960 \times 540$
    \item Labels, containing bounding box coordinates and corresponding query ID
    \item Queries, containing extracted features - in our case distance and bearing relative to the camera pose
\end{itemize}
The queries containing the metric distance and bearing in degrees are normalized by the data loader before being passed to the MLP that computes the embedding. The images are resized to improve the real time capabilities of the network. This decision is based on the computational complexity of the DETR model, where the encoder complexity can be describe as $\mathcal{O}(d^2HW+d(HW)^2)$ and the decoder complexity is $\mathcal{O}(d^2(N+HW)+dNHW)$ \cite{carion_end--end_2020}. One can observe, that the computational complexity is quadratically dependant on $HW$, hence reducing this drastically effects overall inference speed.\\
To train the modified YOLOv7 model, which predicts and additional distance estimate, the training set is slightly adapted to facilitate distance prediction:
\begin{itemize}
    \item Image, resized to $1024 \times 1024$ (retaining aspect ratio)
    \item Label, containing Bounding Box Coordinates (normalized) and the metric distance to the object in meters
\end{itemize}
During training the distance is normalized by a predefined $d_{max}$ value, in our case set to 1000 m. Since the network predicts a normalized value $p_{dist} \in [0,1]$, we apply linear rescaling by multiplying with $d_{max}$ to obtain the metric estimate.\\
All models are trained on a cluster containing eight NVIDIA RTX3090 GPUs. Inference is performed on a GTX 1080 Ti GPU. The transformer fusion model is trained for 100 epochs with a batch size of 8 using the \textit{AdamW} optimizer \cite{Loshchilov2017DecoupledWD} with a learning rate of $1\cdot10^{-4}$ for the transformer and $1\cdot10^{-5}$ for the backbone. Additionally, we use a learning rate scheduler with lr drop at epoch 65. 
% The following loss coefficients are used:
% \begin{itemize}
%     \item $\lambda_{visibility} = 1$
%     \item $\lambda_{box,L1} = 3$
%     \item $\lambda_{box,GIoU} = 7$
% \end{itemize}
Moreover, we apply a variety of augmentation techniques, namely horizontal and vertical flipping w.r.t. image and chart domain and query augmentation by adding a uniform noise over the distance and bearing arguments. 

\subsection{Benchmark models}
To compare the performance of the end-to-end fusion transformer, we provide two different baseline methods to establish correspondence between chart data and detected objects. The first method is a simple ray casting model that uses the unmodified YOLOv7 object detector \cite{wang2023yolov7} to extract bounding box coordinates. Subsequently the ray from the camera position to the object in the ships coordinate system can be computed as:
%\begin{align*}
%    \mathbf{p}_1 &= ^{Ship}\mathbf{T}_{Cam} [x,y,z,1]^T,\\
%    \mathbf{p}_2 &= ^{Ship}\mathbf{T}_{Cam} [0,0,0,1]^T,\\
%    \mathbf{v} &= \mathbf{p}_2 + t \cdot (\mathbf{p}_1-%\mathbf{p}_2), 
%\end{align*}
%with:
%\begin{align*}
%    u &= c_x,\\
%    v &= c_y + h/2,\\
%    x &= \frac{u-u_0}{f_s},\\
%    y &= \frac{v-v_0}{f_s},\\
%    z &= f_l,
%\end{align*}

\[
  % left block
  \begin{aligned}
    \mathbf{p}_1 &= {}^{Ship}\mathbf{T}_{Cam}[x,y,z,1]^T,\\
    \mathbf{p}_2 &= {}^{Ship}\mathbf{T}_{Cam}[0,0,0,1]^T,\\
    \mathbf{v}   &= \mathbf{p}_2 + t\,(\mathbf{p}_1-\mathbf{p}_2),
  \end{aligned}
  \quad
  % the little “with” centered on the math axis
  \vcenter{\hbox{\text{with}}}
  \quad
  % right block
  \begin{aligned}
    u &= c_x,\\
    v &= c_y + \tfrac{h}{2},\\
    x &= \tfrac{u - u_0}{f_s},\\
    y &= \tfrac{v - v_0}{f_s},\\
    z &= f_l
  \end{aligned}
\]

where $u_o$ is the pixel coordinate w.r.t. width of the principal point of the camera and $f_s$, $f_l$ being the camera intrinsics, namely focal length and a scale factor derived from the pixel size.
The transformation matrix $^{Ship}\mathbf{T}_{Cam}$ is derived from IMU recordings, specifically roll, pitch, and yaw, while the translation vector is estimated separately.\\
To approximate the buoy's location, the intersection of the casted ray with the $xy$-plane ($0x+0y+1z = 0$) in the ship's coordinate system is used. The Hungarian matching algorithm \cite{Kuhn1955Hungarian} determines correspondences between predicted objects in world space and chart markers. This method depends on precise IMU measurements, as even minor deviations in camera orientation can lead to significant errors in location estimation.\\
The second benchmark model we propose extends YOLOv7 by incorporating an additional output for distance estimation alongside bounding box coordinates, objectness scores, and class predictions. Previous studies have demonstrated that integrating an extra output node into the detection heads enables accurate distance estimation \cite{vaijgl2020distyolo,kiefer2025approximate}. The estimated distance, combined with the relative bearing between the buoy and the ship, allows for the calculation of the detected object's position in the ship's body coordinate system:
\begin{align*}
    x = cos(\alpha) \cdot dist,\\
    y = sin(\alpha) \cdot dist.
\end{align*}
The bearing is determined as follows:
\begin{align*}
    \alpha = -\arctan \left(\frac{c_x- u_o}{f_sf_l} \right),
\end{align*}
where $c_x$ is the center of the bounding box w.r.t. the $x$ axis (width).
Subsequently, the Hungarian matching algorithm \cite{Kuhn1955Hungarian} is applied to find the optimal correspondence between the predicted object locations and the navigational aids, which are sampled according to the procedure outlined in algorithm Section \ref{cms}. The matching cost is determined by calculating the Euclidean distance between the predicted positions and the ground truth locations of the objects. 
We utilize a Multi Object Tracker (MOT), specifically ByteTrack \cite{zhang2022bytetrack}, to assign unique identifiers to the detected objects and establish correspondence between frames. This enables the calculation of moving averages for the distance estimates, helping to smooth out fluctuations in the predictions and thereby improving their accuracy. We also compute a confidence score for each prediction and its corresponding ground truth, which increases when a consistent match is found across consecutive frames. This is used to perform online calibrations of the focal length parameter $f_s$ and the heading bias $b_h$, which both play a significant role in the computation of the predicted location of objects in the body coordinate system. To ensure an accurate correction, we only compute the bias terms if a high matching confidence is achieved.\\
In summary, this both benchmark models depend on several manually crafted post-processing steps to establish correspondences, which stands in sharp contrast to the proposed transformer network that directly predicts the matched objects along with their bounding box estimates.

\subsection{Experimental Results}
\begin{figure}
    \centering
    \includegraphics[width=0.99\linewidth]{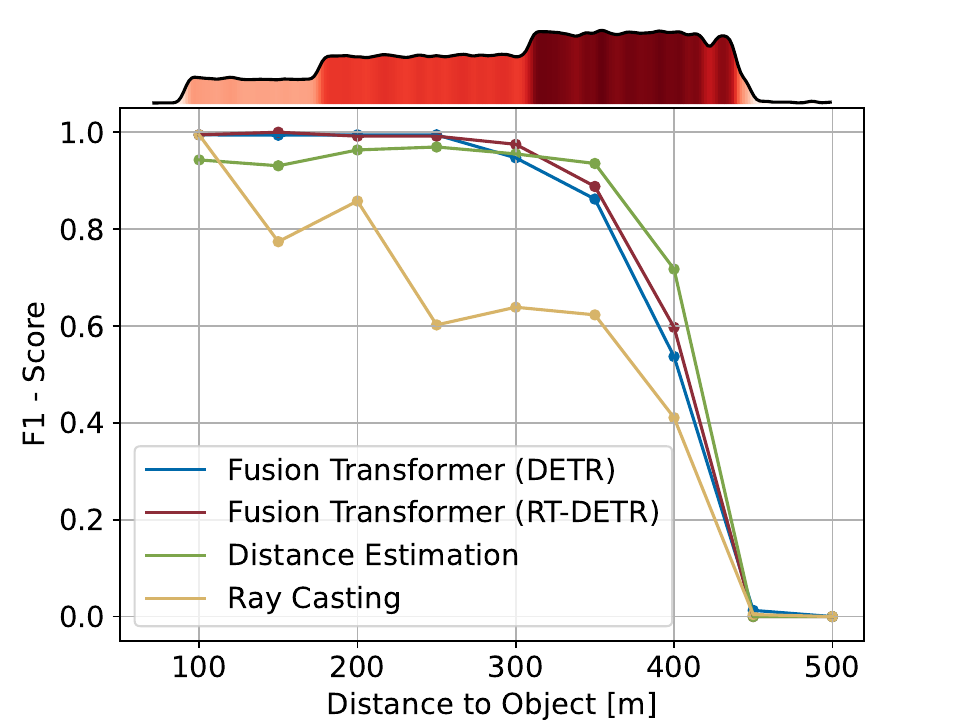}
    \caption{Visualization of the F1-Score as a function of the Ground Truth Distance to the labeled objects, with the distribution of the labeled objects shown at the top.}
    \label{fig:f1_over_dist}
\end{figure}

\begin{table*}[h!]  % Use table* instead of table
    \centering
    \caption{Inference results of the different methods on the test sequence. The transformer based fusion networks outperform the benchmark models in terms of accuracy and inference speed. The latency is measured on a GTX 1080 Ti GPU.\\}
    \resizebox{\textwidth}{!}{  % Rescale to full width
        \begin{tabular}{lcccccccc}
            \toprule
             Method & Resolution & Parameters & Precision & Recall & F1-Score & Mean-IoU & Latency [ms] & FPS \\
            \hline
            \hline
            Raycasting \& Matching & $1024 \times 1024$ & 36.4M & 0.440 & 0.675 & 0.533 & 0.528 & 57 & 17.5\\
            Distance Estimation \& Matching & $1024 \times 1024$ & 36.4M & 0.826 & \textbf{0.890} & 0.857 & 0.727 & 56 & 18\\
            Fusion Transformer based on DETR \cite{carion_end--end_2020} & $960 \times 540$ & 41.3M & 0.859 & 0.869 & 0.864 & 0.729 & 32 & 31.3\\
            Fusion Transformer based on RT-DETR \cite{lv2023detrs} & $960 \times 540$ & 49.1M & \textbf{0.905} & 0.881 & \textbf{0.893} & \textbf{0.744} & 44 & 22.8\\
            \bottomrule
        \end{tabular}
    }
    \label{tab:benchmark-results}
\end{table*}
Table \ref{tab:benchmark-results} compares the fusion transformer networks, based on DETR \cite{carion_end--end_2020} and RT-DETR \cite{lv2023detrs}, with the benchmark models. The Fusion Transformer derived from DETR only operates on an activation map that is extracted at one specific layer of the backbone, thus limiting the performance to detect both large scale and small scale objects. We also adopt the architectural modifications to the RT-DETR model in a similar manner, by replacing the decoder inputs with the sampled buoy queries. This model incorporates activation maps at different scales and uses \textit{Deformable Attention} \cite{zhu_deformable_2021} to ensure real-time capability.\\
The results indicate that the Fusion Transformers outperform both the raycasting approach and the model that jointly predicts distance and bounding box coordinates. Their inference latency is lower than that of YOLOv7-based benchmark methods, as the number of input decoder queries, $N$, is typically small. The test sequence contains 4.6 queries on average. However, this number varies depending on the number of surrounding buoys. If the count exceeds a certain threshold, limiting the sequence length may help maintain real-time performance, similar to setting a maximum context length in large language models. In contrast, DETR and RT-DETR utilize between 100 and 300 object queries, which increases latency. The reason for the higher inference latency of the model that is derived from RT-DETR can be attributed to the fact that in order to achieve satisfactory matching accuracy, the number of points to which each query attends to in the decoder has to be increased. In our case, we settle for 64. Since we only operate on few queries, a small number of sampling points can result in missing entire objects, thus decreasing the recall metric. This is not the case for the base DETR model which attends to all embeddings of the encoder. In Figure \ref{fig:f1_over_dist} the F1-Score over the ground truth distance to the objects occuring in the frame is shown. Notably, the YOLOv7 based Distance Estimator works best for distant objects, since the DETR models typically have smaller $AP_s$ values.

\subsection{Ablation Study on Decoder Embedding}
Table \ref{tab:abls_embedding} presents the inference results for the fusion transformer based on DETR, comparing two approaches for obtaining input embeddings for the decoder. In the first approach, learned embeddings are used and concatenated to form vectors of size $d_{model}$, while the second approach employs a two-layer MLP. Since the object selection process provides continuous features for the filtered buoys, discretization is required to use learned embeddings.
To achieve this, the distance values are first clamped to the range $[0,1000]$ and then divided by 25 using integer division, producing 41 possible embeddings. Similarly, bearing angles are scaled to $[0,1000]$ and divided by 12.5, resulting in 81 embeddings. Both embeddings have size $d_{model}/2$. Applying concatenation results in the input embedding $\mathbf{v} \in \mathbb{R}^{d_{model}}$. In contrast, the MLP does not require discretization, allowing direct input of the normalized distance $d \in [0,1]$ and bearing $b \in [-1, 1]$.
\begin{table}[]
    \centering
    \caption{Ablation study results comparing learned embeddings as decoder input with embeddings generated by an MLP. It reports precision, recall F1-score and mean intersection over union.\\}
    \begin{tabular}{l|cccc}
        \toprule
         Embedding  &  P & R & F1 & Mean-IoU\\
         \hline
         \hline
         Learned & 0.804 & 0.774 & 0.788 & 0.666\\
         MLP & 0.859 & 0.869 & 0.864 & 0.729\\
         \bottomrule
    \end{tabular}
    \label{tab:abls_embedding}
\end{table}

\subsection{Ablation Study on Number of Sampling Points}
We adapt the RT-DETR model \cite{lv2023detrs} by incorporating the described architectural modifications, replacing the decoder’s object queries with dynamically sampled chart markers. As a result, the decoder operates with a significantly reduced number of queries, averaging between 4 and 5, compared to the usual 100 to 300 queries in RT-DETR, which are initialized using the \textit{top-k} encoder embeddings. During deformable attention, each query predicts a set of sampling points, allowing attention to be performed on a reduced subset of elements. This approach minimizes latency and prevents the computational cost from scaling quadratically with sequence length \cite{zhu_deformable_2021}. However, this means that a query is not attending to all encoder embeddings and might miss relevant ones. The number of sample points is usually set quite low, by default to 4. This problem can be migitated by a high number of object queries, however this is not the case for our approach. Hence we strive to improve the performance of the model by increasing the number of sampling points. The results are detailed in Table \ref{tab:abls_sampling_points}.

\begin{table}[]
    \centering
    \caption{Ablation study on the number of sampling points ($\Delta p$) employed in the decoder for the RT-DETR based transformer.\\}
    \begin{tabular}{c|ccccc}
        \toprule
         \begin{tabular}{@{}c@{}}\#Sampling \\ Points\end{tabular} &  P & R & F1 & mIoU & FPS\\
         \hline
         \hline
         4 & 0.799 & 0.883 & 0.839 & 0.722 & 25.95\\
         8 & 0.908 & 0.811 & 0.857 & 0.727 & 25.54\\
         16 & 0.861 & 0.876 & 0.868 & 0.723 & 24.06\\
         32 & 0.852 & 0.894 & 0.872 & 0.734 & 23.36\\
         64 & 0.905 & 0.881 & 0.893 & 0.744 & 22.83\\
         \bottomrule
    \end{tabular}
    \label{tab:abls_sampling_points}
\end{table}

% \subsection{Ablation Study on the Selected Features}
% As stated in section \ref{sec:model_architecture}, the number of features per object used to generate the decoder input embeddings is flexible. In our standard approach, we incorporate the bearing and distance between the camera and buoys. In this study, we expand the feature set by including visual attributes such as color, shape, and object class. These additional features, along with buoy positions, are sourced from the NOAA database \cite{noaa_db}.

%% file: sec/5_discussion.tex
\begin{figure*}[ht!]
    \centering
    \begin{subfigure}[t]{0.5\linewidth}
        \centering
        \includegraphics[height=4cm]{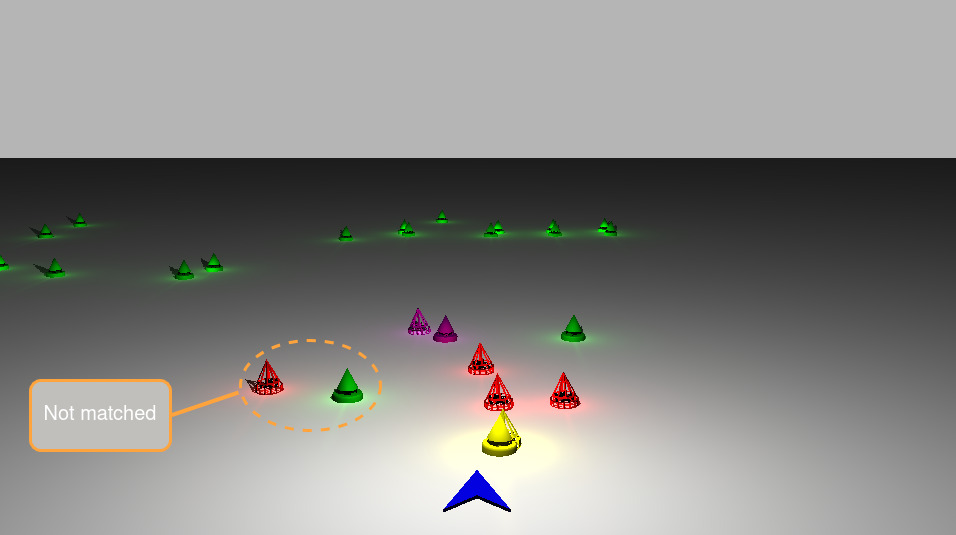}
        \caption{Association of Predictions and Ground Truth}
    \end{subfigure}%
    ~ 
    \begin{subfigure}[t]{0.5\linewidth}
        \centering
        \includegraphics[height=4cm]{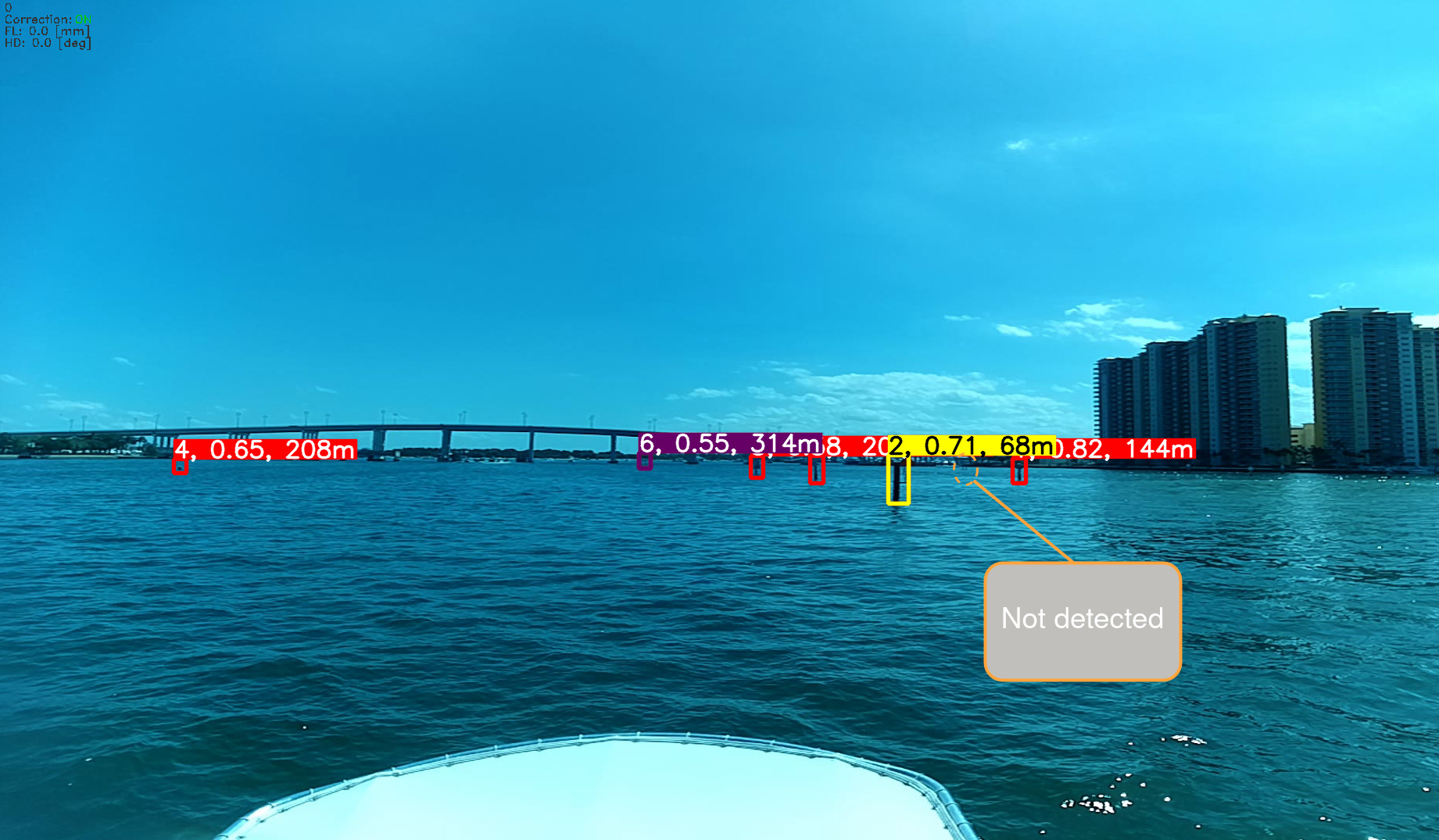}
        \caption{Detection}
    \end{subfigure}
    \caption{Associations and detections of the YOLOv7 based distance estimation model. Wireframe buoys in (a) are predictions, transformed to the ships CS. Matched pairs have the same color. Unmatched predictions are red, unmatched chart markers green. The yellow and purple associations are correct. The three red buoys in the center are not marked on the chart, hence they are are also correctly not matched to any green ground truth buoy. The red buoy on the left hand side however should be matched to the green marker next to it, since this is the correct corresponding ground truth marker.}
    \label{fig:yolov7_disc}
\end{figure*}

\begin{figure*}[ht!]
    \centering
    \begin{subfigure}[t]{0.5\linewidth}
        \centering
        \includegraphics[height=4cm]{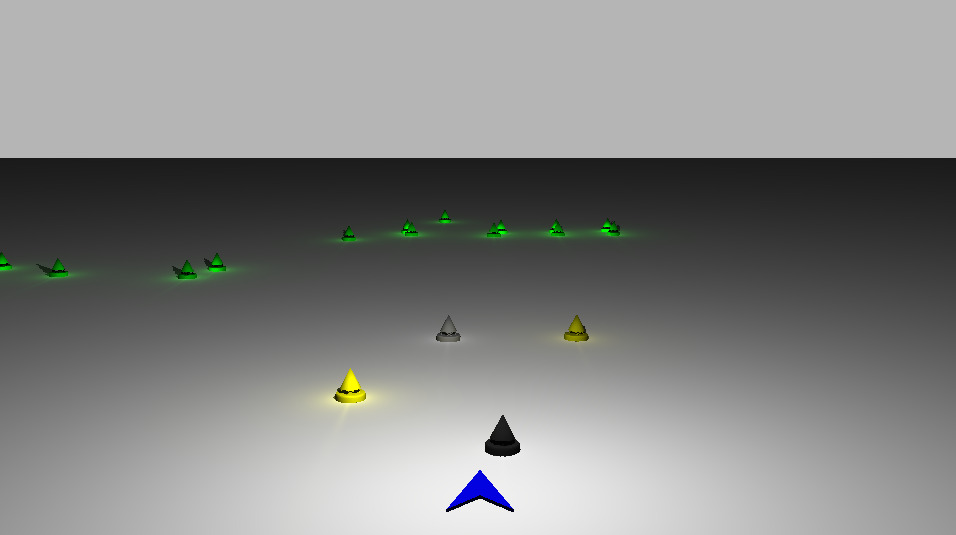}
        \caption{Association}
    \end{subfigure}%
    ~ 
    \begin{subfigure}[t]{0.5\linewidth}
        \centering
        \includegraphics[height=4cm]{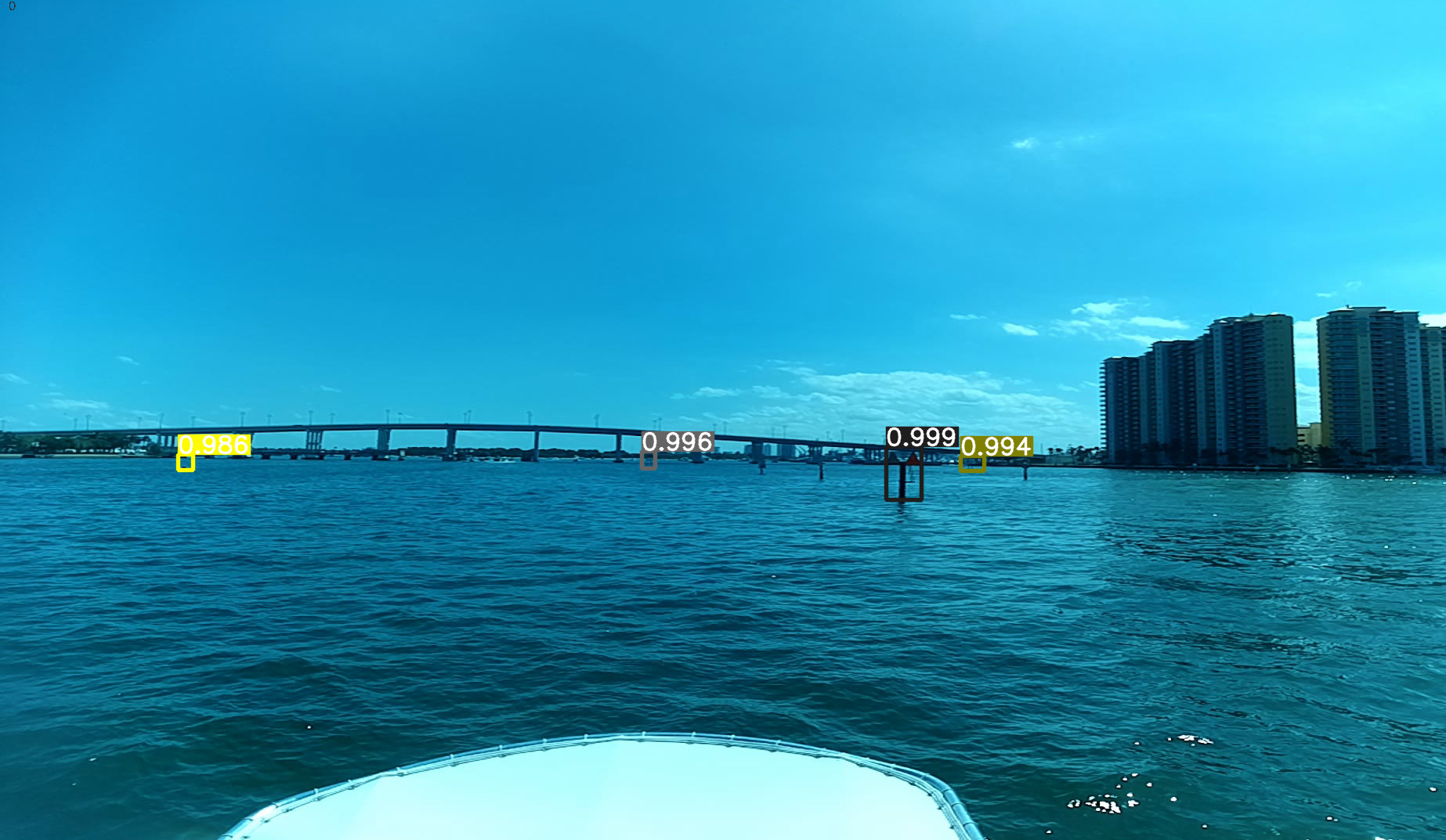}
        \caption{Detection}
    \end{subfigure}
    \caption{Associations and detections of the Fusion Transformer based on DETR. The computed matching is correct. It manages to detect the small distant buoy on the right hand side and matches it correctly. The visible buoys that do not have corresponding chart markers are not matched, which is desired.}
    \label{fig:transformer_disc}
\end{figure*}

\section{Discussion, Limitations, Conclusions}
\label{sec_discussion}
The proposed Fusion Transformers are capable of achieving superior results by performing end to end inference compared with the two baseline models, which require many postprocessing steps to compute an accurate matching. Notably, the best recall score is achieved by the distance estimation network, which can likely be attributed to the inferiority of detecting small objects of DETR based networks, as described in \cite{lv2023detrs, zhu_deformable_2021}. Qualitative analysis of the proposed Fusion Transformer shows that it especially excels in scenarios, where a precise matching is a difficult task. 
Consider Figure \ref{fig:yolov7_disc}, where many buoys are visible in the image domain, however only a few are mapped on the chart. The distance estimation and matching approach is not able to establish the correct correspondence, since strict matching filtering has to be employed to avoid visible but not mapped buoys to be associated with a chart marker that is not matched. However, this leads to unmatched prediction-ground truth pairs, where the spatial estimate is incorrect. 
The Fusion Transformer is able to compute a correct correspondence for this scene, see Figure \ref{fig:transformer_disc}.
Additionally, we observe that all the algorithms have difficulty accurately matching detected buoys to the corresponding chart markers when the chart markers are incorrectly mapped, such as in the case of significant deviations between a mapped position of a navigational aid and its actual position. This issue is especially prominent at close ranges, where the incorrectly mapped position causes a substantial difference between the observed and computed bearing to the buoy.

In summary, our method is robust and capable of establishing correct associations in a variety of real world cases. Future work could extend this approach to other domains by modifying the object sampling process and adapting query embeddings.
Using 2D polar coordinates as embeddings allows flexible adaptation. Unlike DETR, we achieve faster convergence by using strong priors for object locations, replacing object queries with sampled buoy queries, and eliminating the need for Hungarian matching. These adaptations streamline training, and the proposed Fusion Transformer outperforms benchmark models without requiring post-processing.

%% file: main.bbl
\begin{thebibliography}{44}
\providecommand{\natexlab}[1]{#1}
\providecommand{\url}[1]{\texttt{#1}}
\expandafter\ifx\csname urlstyle\endcsname\relax
  \providecommand{\doi}[1]{doi: #1}\else
  \providecommand{\doi}{doi: \begingroup \urlstyle{rm}\Url}\fi

\bibitem[Aitken et~al.(2021)Aitken, Evans, Worley, Edwards, Zhang, Dodd, Mihaylova, and Anderson]{aitken2021simultaneous}
Jonathan~M Aitken, Mathew~H Evans, Rob Worley, Sarah Edwards, Rui Zhang, Tony Dodd, Lyudmila Mihaylova, and Sean~R Anderson.
\newblock Simultaneous localization and mapping for inspection robots in water and sewer pipe networks: A review.
\newblock \emph{IEEE access}, 9:\penalty0 140173--140198, 2021.

\bibitem[Ardeshir et~al.(2014)Ardeshir, Zamir, Torroella, and Shah]{ardeshir2014gis}
Shervin Ardeshir, Amir~Roshan Zamir, Alejandro Torroella, and Mubarak Shah.
\newblock Gis-assisted object detection and geospatial localization.
\newblock In \emph{Computer Vision--ECCV 2014: 13th European Conference, Zurich, Switzerland, September 6-12, 2014, Proceedings, Part VI 13}, pages 602--617. Springer, 2014.

\bibitem[Bae et~al.(2015)Bae, Kim, Kim, Choi, and Ahn]{bae2015development}
Ho~Seuk Bae, Wan-Jin Kim, Woo-Shik Kim, Sang-Moon Choi, and Jin-Hyeong Ahn.
\newblock Development of the sonar system for an unmanned surface vehicle.
\newblock \emph{Journal of the Korea Institute of Military Science and Technology}, 18\penalty0 (4):\penalty0 358--368, 2015.

\bibitem[Biljecki and Ito(2021)]{biljecki2021street}
Filip Biljecki and Koichi Ito.
\newblock Street view imagery in urban analytics and gis: A review.
\newblock \emph{Landscape and Urban Planning}, 215:\penalty0 104217, 2021.

\bibitem[Bovcon et~al.(2018)Bovcon, Per{\v{s}}, Kristan, et~al.]{bovcon2018stereo}
Borja Bovcon, Janez Per{\v{s}}, Matej Kristan, et~al.
\newblock Stereo obstacle detection for unmanned surface vehicles by imu-assisted semantic segmentation.
\newblock \emph{Robotics and Autonomous Systems}, 104:\penalty0 1--13, 2018.

\bibitem[Bovcon et~al.(2019)Bovcon, Muhovi{\v{c}}, Per{\v{s}}, and Kristan]{bovcon2019mastr1325}
Borja Bovcon, Jon Muhovi{\v{c}}, Janez Per{\v{s}}, and Matej Kristan.
\newblock The mastr1325 dataset for training deep usv obstacle detection models.
\newblock In \emph{2019 IEEE/RSJ International Conference on Intelligent Robots and Systems (IROS)}, pages 3431--3438. IEEE, 2019.

\bibitem[Bovcon et~al.(2021)Bovcon, Muhovi{\v{c}}, Vranac, Mozeti{\v{c}}, Per{\v{s}}, and Kristan]{bovcon2021mods}
Borja Bovcon, Jon Muhovi{\v{c}}, Du{\v{s}}ko Vranac, Dean Mozeti{\v{c}}, Janez Per{\v{s}}, and Matej Kristan.
\newblock Mods—a usv-oriented object detection and obstacle segmentation benchmark.
\newblock \emph{IEEE Transactions on Intelligent Transportation Systems}, 23\penalty0 (8):\penalty0 13403--13418, 2021.

\bibitem[Carion et~al.(2020)Carion, Massa, Synnaeve, Usunier, Kirillov, and Zagoruyko]{carion_end--end_2020}
Nicolas Carion, Francisco Massa, Gabriel Synnaeve, Nicolas Usunier, Alexander Kirillov, and Sergey Zagoruyko.
\newblock End-to-end object detection with transformers.
\newblock In \emph{Computer Vision – ECCV 2020: 16th European Conference, Glasgow, UK, August 23–28, 2020, Proceedings, Part I}, page 213–229, Berlin, Heidelberg, 2020. Springer-Verlag.

\bibitem[Chen et~al.(2023)Chen, Chen, Wang, Zhang, Yao, Feng, Han, Ding, Zeng, and Wang]{Chen_2023_ICCV}
Qiang Chen, Xiaokang Chen, Jian Wang, Shan Zhang, Kun Yao, Haocheng Feng, Junyu Han, Errui Ding, Gang Zeng, and Jingdong Wang.
\newblock Group detr: Fast detr training with group-wise one-to-many assignment.
\newblock In \emph{Proceedings of the IEEE/CVF International Conference on Computer Vision (ICCV)}, pages 6633--6642, 2023.

\bibitem[Czaplewski and {\'S}wierczy{\'n}ski(2021)]{czaplewski2021method}
Krzysztof Czaplewski and S{\l}awomir {\'S}wierczy{\'n}ski.
\newblock A method of increasing the accuracy of radar distance measurement in vts systems for vessels with very large dimensions.
\newblock \emph{Remote Sensing}, 13\penalty0 (16):\penalty0 3066, 2021.

\bibitem[Dai et~al.(2021{\natexlab{a}})Dai, Chen, Yang, Zhang, Yuan, and Zhang]{Dai_2021_ICCV}
Xiyang Dai, Yinpeng Chen, Jianwei Yang, Pengchuan Zhang, Lu Yuan, and Lei Zhang.
\newblock Dynamic detr: End-to-end object detection with dynamic attention.
\newblock In \emph{Proceedings of the IEEE/CVF International Conference on Computer Vision (ICCV)}, pages 2988--2997, 2021{\natexlab{a}}.

\bibitem[Dai et~al.(2021{\natexlab{b}})Dai, Cai, Lin, and Chen]{Dai_2021_CVPR}
Zhigang Dai, Bolun Cai, Yugeng Lin, and Junying Chen.
\newblock Up-detr: Unsupervised pre-training for object detection with transformers.
\newblock In \emph{Proceedings of the IEEE/CVF Conference on Computer Vision and Pattern Recognition (CVPR)}, pages 1601--1610, 2021{\natexlab{b}}.

\bibitem[GeoPlatform()]{noaa_db}
NOAA GeoPlatform.
\newblock Marine {C}adastre - {A}ids to {N}avigation.

\bibitem[Gladstone et~al.(2016)Gladstone, Moshe, Barel, and Shenhav]{gladstone2016distance}
Ran Gladstone, Yair Moshe, Avihai Barel, and Elior Shenhav.
\newblock Distance estimation for marine vehicles using a monocular video camera.
\newblock In \emph{2016 24th European Signal Processing Conference (EUSIPCO)}, pages 2405--2409. IEEE, 2016.

\bibitem[G\"ulsoylu et~al.(2024)G\"ulsoylu, Koch, Yildiz, Constapel, and Kelm]{Gulsoylu_2024_WACV}
Emre G\"ulsoylu, Paul Koch, Mert Yildiz, Manfred Constapel, and Andr\'e~Peter Kelm.
\newblock Image and ais data fusion technique for maritime computer vision applications.
\newblock In \emph{Proceedings of the IEEE/CVF Winter Conference on Applications of Computer Vision (WACV) Workshops}, pages 859--868, 2024.

\bibitem[Gundogdu et~al.(2017)Gundogdu, Solmaz, Y{\"u}cesoy, and Koc]{gundogdu2017marvel}
Erhan Gundogdu, Berkan Solmaz, Veysel Y{\"u}cesoy, and Aykut Koc.
\newblock Marvel: A large-scale image dataset for maritime vessels.
\newblock In \emph{Computer Vision--ACCV 2016: 13th Asian Conference on Computer Vision, Taipei, Taiwan, November 20-24, 2016, Revised Selected Papers, Part V 13}, pages 165--180. Springer, 2017.

\bibitem[Kiefer and Zell(2024)]{kiefer2024real}
Benjamin Kiefer and Andreas Zell.
\newblock Real-time horizon locking on unmanned surface vehicles.
\newblock In \emph{2024 IEEE/RSJ International Conference on Intelligent Robots and Systems (IROS)}, pages 1214--1221. IEEE, 2024.

\bibitem[Kiefer et~al.(2023{\natexlab{a}})Kiefer, H{\"o}fer, and Zell]{kiefer2023stable}
Benjamin Kiefer, Timon H{\"o}fer, and Andreas Zell.
\newblock Stable yaw estimation of boats from the viewpoint of uavs and usvs.
\newblock In \emph{2023 European Conference on Mobile Robots (ECMR)}, pages 1--6. IEEE, 2023{\natexlab{a}}.

\bibitem[Kiefer et~al.(2023{\natexlab{b}})Kiefer, Kristan, Per{\v{s}}, {\v{Z}}ust, Poiesi, Andrade, Bernardino, Dawkins, Raitoharju, Quan, et~al.]{kiefer20231st}
Benjamin Kiefer, Matej Kristan, Janez Per{\v{s}}, Lojze {\v{Z}}ust, Fabio Poiesi, Fabio Andrade, Alexandre Bernardino, Matthew Dawkins, Jenni Raitoharju, Yitong Quan, et~al.
\newblock 1st workshop on maritime computer vision (macvi) 2023: Challenge results.
\newblock In \emph{Proceedings of the IEEE/CVF Winter Conference on Applications of Computer Vision}, pages 265--302, 2023{\natexlab{b}}.

\bibitem[Kiefer et~al.(2023{\natexlab{c}})Kiefer, Quan, and Zell]{kiefer2023memory}
Benjamin Kiefer, Yitong Quan, and Andreas Zell.
\newblock Memory maps for video object detection and tracking on uavs.
\newblock In \emph{2023 IEEE/RSJ International Conference on Intelligent Robots and Systems (IROS)}, pages 3040--3047. IEEE, 2023{\natexlab{c}}.

\bibitem[Kiefer et~al.(2023{\natexlab{d}})Kiefer, {\v{Z}}ust, Kristan, Per{\v{s}}, Ter{\v{s}}ek, Wiliem, Messmer, Yang, Huang, Jiang, et~al.]{kiefer20232nd}
Benjamin Kiefer, Lojze {\v{Z}}ust, Matej Kristan, Janez Per{\v{s}}, Matija Ter{\v{s}}ek, Arnold Wiliem, Martin Messmer, Cheng-Yen Yang, Hsiang-Wei Huang, Zhongyu Jiang, et~al.
\newblock The 2nd workshop on maritime computer vision (macvi) 2024.
\newblock \emph{arXiv preprint arXiv:2311.14762}, 2023{\natexlab{d}}.

\bibitem[Kiefer et~al.(2025{\natexlab{a}})Kiefer, Quan, and Zell]{kiefer2025approximate}
Benjamin Kiefer, Yitong Quan, and Andreas Zell.
\newblock Approximate supervised object distance estimation on unmanned surface vehicles.
\newblock \emph{arXiv preprint arXiv:2501.05567}, 2025{\natexlab{a}}.

\bibitem[Kiefer et~al.(2025{\natexlab{b}})Kiefer, Zust, Kristan, Pers, Tersek, Mudenagudi, Desai, Wiliem, Kreis, Akalwadi, et~al.]{kiefer20253rd}
Benjamin Kiefer, Lojze Zust, Matej Kristan, Janez Pers, Matija Tersek, Uma Mudenagudi, Chaitra Desai, Arnold Wiliem, Marten Kreis, Nikhil Akalwadi, et~al.
\newblock 3rd workshop on maritime computer vision (macvi) 2025: Challenge results.
\newblock In \emph{Proceedings of the Winter Conference on Applications of Computer Vision}, pages 1542--1569, 2025{\natexlab{b}}.

\bibitem[Kristan et~al.(2015)Kristan, Kenk, Kova{\v{c}}i{\v{c}}, and Per{\v{s}}]{kristan2015fast}
Matej Kristan, Vildana~Suli{\'c} Kenk, Stanislav Kova{\v{c}}i{\v{c}}, and Janez Per{\v{s}}.
\newblock Fast image-based obstacle detection from unmanned surface vehicles.
\newblock \emph{IEEE transactions on cybernetics}, 46\penalty0 (3):\penalty0 641--654, 2015.

\bibitem[Krylov et~al.(2018)Krylov, Kenny, and Dahyot]{krylov2018automatic}
Vladimir~A Krylov, Eamonn Kenny, and Rozenn Dahyot.
\newblock Automatic discovery and geotagging of objects from street view imagery.
\newblock \emph{Remote Sensing}, 10\penalty0 (5):\penalty0 661, 2018.

\bibitem[Kuhn(1955)]{Kuhn1955Hungarian}
Harold~W. Kuhn.
\newblock {The Hungarian Method for the Assignment Problem}.
\newblock \emph{Naval Research Logistics Quarterly}, 2\penalty0 (1--2):\penalty0 83--97, 1955.

\bibitem[Li et~al.(2022)Li, Zhang, Liu, Guo, Ni, and Zhang]{li2022dn}
Feng Li, Hao Zhang, Shilong Liu, Jian Guo, Lionel~M Ni, and Lei Zhang.
\newblock Dn-detr: Accelerate detr training by introducing query denoising.
\newblock In \emph{Proceedings of the IEEE/CVF Conference on Computer Vision and Pattern Recognition}, pages 13619--13627, 2022.

\bibitem[Li and Ibanez-Guzman(2020)]{li2020lidar}
You Li and Javier Ibanez-Guzman.
\newblock Lidar for autonomous driving: The principles, challenges, and trends for automotive lidar and perception systems.
\newblock \emph{IEEE Signal Processing Magazine}, 37\penalty0 (4):\penalty0 50--61, 2020.

\bibitem[Liu et~al.(2022)Liu, Li, Zhang, Yang, Qi, Su, Zhu, and Zhang]{liu2022dabdetr}
Shilong Liu, Feng Li, Hao Zhang, Xiao Yang, Xianbiao Qi, Hang Su, Jun Zhu, and Lei Zhang.
\newblock {DAB}-{DETR}: Dynamic anchor boxes are better queries for {DETR}.
\newblock In \emph{International Conference on Learning Representations}, 2022.

\bibitem[Loshchilov and Hutter(2017)]{Loshchilov2017DecoupledWD}
Ilya Loshchilov and Frank Hutter.
\newblock Decoupled weight decay regularization.
\newblock In \emph{International Conference on Learning Representations}, 2017.

\bibitem[Meng et~al.(2021)Meng, Chen, Fan, Zeng, Li, Yuan, Sun, and Wang]{Meng_2021_ICCV}
Depu Meng, Xiaokang Chen, Zejia Fan, Gang Zeng, Houqiang Li, Yuhui Yuan, Lei Sun, and Jingdong Wang.
\newblock Conditional detr for fast training convergence.
\newblock In \emph{Proceedings of the IEEE/CVF International Conference on Computer Vision (ICCV)}, pages 3651--3660, 2021.

\bibitem[Moosbauer et~al.(2019)Moosbauer, Konig, Jakel, and Teutsch]{moosbauer2019benchmark}
Sebastian Moosbauer, Daniel Konig, Jens Jakel, and Michael Teutsch.
\newblock A benchmark for deep learning based object detection in maritime environments.
\newblock In \emph{Proceedings of the IEEE/CVF conference on computer vision and pattern recognition workshops}, pages 0--0, 2019.

\bibitem[Ruiz-Ponce et~al.(2023)Ruiz-Ponce, Ortiz-Perez, Garcia-Rodriguez, and Kiefer]{ruiz2023poseidon}
Pablo Ruiz-Ponce, David Ortiz-Perez, Jose Garcia-Rodriguez, and Benjamin Kiefer.
\newblock Poseidon: A data augmentation tool for small object detection datasets in maritime environments.
\newblock \emph{Sensors}, 23\penalty0 (7):\penalty0 3691, 2023.

\bibitem[Thombre et~al.(2020)Thombre, Zhao, Ramm-Schmidt, Garcia, Malkam{\"a}ki, Nikolskiy, Hammarberg, Nuortie, Bhuiyan, S{\"a}rkk{\"a}, et~al.]{thombre2020sensors}
Sarang Thombre, Zheng Zhao, Henrik Ramm-Schmidt, Jose M~Vallet Garcia, Tuomo Malkam{\"a}ki, Sergey Nikolskiy, Toni Hammarberg, Hiski Nuortie, M~Zahidul~H Bhuiyan, Simo S{\"a}rkk{\"a}, et~al.
\newblock Sensors and ai techniques for situational awareness in autonomous ships: A review.
\newblock \emph{IEEE transactions on intelligent transportation systems}, 23\penalty0 (1):\penalty0 64--83, 2020.

\bibitem[Tu et~al.(2017)Tu, Zhang, Rachmawati, Rajabally, and Huang]{tu2017exploiting}
Enmei Tu, Guanghao Zhang, Lily Rachmawati, Eshan Rajabally, and Guang-Bin Huang.
\newblock Exploiting ais data for intelligent maritime navigation: A comprehensive survey from data to methodology.
\newblock \emph{IEEE Transactions on Intelligent Transportation Systems}, 19\penalty0 (5):\penalty0 1559--1582, 2017.

\bibitem[Vajgl et~al.(2022)Vajgl, Hurtik, and Nejezchleba]{vaijgl2020distyolo}
Marek Vajgl, Petr Hurtik, and Tomáš Nejezchleba.
\newblock Dist-yolo: Fast object detection with distance estimation.
\newblock \emph{Applied Sciences}, 12\penalty0 (3), 2022.

\bibitem[Varga et~al.(2022)Varga, Kiefer, Messmer, and Zell]{varga2022seadronessee}
Leon~Amadeus Varga, Benjamin Kiefer, Martin Messmer, and Andreas Zell.
\newblock Seadronessee: A maritime benchmark for detecting humans in open water.
\newblock In \emph{Proceedings of the IEEE/CVF winter conference on applications of computer vision}, pages 2260--2270, 2022.

\bibitem[Vaswani et~al.(2017)Vaswani, Shazeer, Parmar, Uszkoreit, Jones, Gomez, Kaiser, and Polosukhin]{10.5555/3295222.3295349}
Ashish Vaswani, Noam Shazeer, Niki Parmar, Jakob Uszkoreit, Llion Jones, Aidan~N. Gomez, \L{}ukasz Kaiser, and Illia Polosukhin.
\newblock Attention is all you need.
\newblock In \emph{Proceedings of the 31st International Conference on Neural Information Processing Systems}, page 6000–6010, Red Hook, NY, USA, 2017. Curran Associates Inc.

\bibitem[Wang et~al.(2023)Wang, Bochkovskiy, and Liao]{wang2023yolov7}
Chien-Yao Wang, Alexey Bochkovskiy, and Hong-Yuan~Mark Liao.
\newblock Yolov7: Trainable bag-of-freebies sets new state-of-the-art for real-time object detectors.
\newblock In \emph{2023 IEEE/CVF Conference on Computer Vision and Pattern Recognition (CVPR)}, pages 7464--7475, 2023.

\bibitem[Yao et~al.(2021)Yao, Ai, Li, and Zhang]{yao2021efficientdetrimprovingendtoend}
Zhuyu Yao, Jiangbo Ai, Boxun Li, and Chi Zhang.
\newblock Efficient detr: Improving end-to-end object detector with dense prior, 2021.

\bibitem[Zhang et~al.(2022{\natexlab{a}})Zhang, Li, Liu, Zhang, Su, Zhu, Ni, and Shum]{zhang2022dino}
Hao Zhang, Feng Li, Shilong Liu, Lei Zhang, Hang Su, Jun Zhu, Lionel~M. Ni, and Heung-Yeung Shum.
\newblock Dino: Detr with improved denoising anchor boxes for end-to-end object detection, 2022{\natexlab{a}}.

\bibitem[Zhang et~al.(2022{\natexlab{b}})Zhang, Sun, Jiang, Yu, Weng, Yuan, Luo, Liu, and Wang]{zhang2022bytetrack}
Yifu Zhang, Peize Sun, Yi Jiang, Dongdong Yu, Fucheng Weng, Zehuan Yuan, Ping Luo, Wenyu Liu, and Xinggang Wang.
\newblock Bytetrack: Multi-object tracking by associating every detection box.
\newblock 2022{\natexlab{b}}.

\bibitem[Zhao et~al.(2023)Zhao, Lv, Xu, Wei, Wang, Dang, Liu, and Chen]{lv2023detrs}
Yian Zhao, Wenyu Lv, Shangliang Xu, Jinman Wei, Guanzhong Wang, Qingqing Dang, Yi Liu, and Jie Chen.
\newblock Detrs beat yolos on real-time object detection, 2023.

\bibitem[Zhu et~al.(2021)Zhu, Su, Lu, Li, Wang, and Dai]{zhu_deformable_2021}
Xizhou Zhu, Weijie Su, Lewei Lu, Bin Li, Xiaogang Wang, and Jifeng Dai.
\newblock Deformable {DETR}: {Deformable} {Transformers} for {End}-to-{End} {Object} {Detection}.
\newblock In \emph{International Conference on Learning Representations}, 2021.

\end{thebibliography}
